# Mind the Gap: A Dual Knowledge Graph Framework for Unified Multi-task User Intent Inference

***Published in the PACIS 2026 Proceedings as a Completed Research Paper. AIS eLibrary:***

Tzu-Cheng Peng - National Taiwan University, Taiwan
Chien Chin Chen - National Taiwan University, Taiwan
Chih-Hao Ku - University of North Texas, USA
Yung-Chun Chang - Taipei Medical University, Taiwan

## Introduction

In the contemporary digital landscape, the modern traveler is confronted with a paradox of choice. Platforms such as TripAdvisor, Booking.com, and Google Reviews have democratized information, creating an unprecedented repository of user-generated content (UGC) (Chang et al., 2019; Ma et al., 2024; Zhao et al., 2024). A single hotel may have thousands of reviews, each a rich narrative of personal experience, replete with ratings, photos, and detailed commentary. While this wealth of data promises to empower consumers, it often leads to information overload and decision paralysis. The critical challenge for next-generation intelligent tourism systems, therefore, is not merely to aggregate this data, but to synthesize it, to look beyond explicit ratings and uncover the latent, underlying intent of the user (Guo et al., 2016). This intent, a complex tapestry woven from a traveler's goals, unstated preferences, contextual constraints, and past experiences is the true key to unlocking genuinely personalized recommendations and services.

The academic and commercial pursuit of understanding user intent has evolved significantly with the advent of advanced Natural Language Understanding (NLU) techniques (Mehra, 2022; Ouaddi et al., 2025). The goal is to reverse-engineer a user's review to answer a fundamental question: "What was this person truly looking for?" Answering this requires moving past coarse sentiment analysis to a more granular, multi-faceted interpretation (Peng et al., 2026). For instance, a review praising a hotel's vibrant nightlife and another lauding its peaceful, quiet environment may both carry a positive sentiment, but they reflect diametrically opposed user intents. The former likely belongs to a leisure traveler or a group of friends seeking entertainment, while the latter may be a business traveler needing rest or a family with young children. Accurately distinguishing between these intents is the cornerstone of effective personalization, enabling a system to recommend not just a good hotel, but the right hotel for a specific user in a specific context.

Recognizing the limitations of static fine-tuning, its high computational cost and inability to easily update internalized knowledge, another prominent paradigm, Retrieval-Augmented Generation (RAG), has emerged (Lewis et al., 2020). RAG-based systems ground Large Language Models (LLMs) with external knowledge at inference time by utilizing a retriever-generator architecture (Karpukhin et al., 2020). This approach offers greater flexibility than traditional fine-tuning, as it allows for real-time information updates and provides a verifiable source of truth for the model's outputs. However, standard RAG implementations are often hobbled by their structural blindness. They typically retrieve information based on lexical or semantic similarity, matching keywords or phrases from the user's query to a vector database. This approach is effective for fact-based question answering but falls short in domains requiring deep structural reasoning. For example, a user review mentioning a successful business trip has a structural, not merely lexical, relationship with concepts like "fast Wi-Fi," "24-hour check-in," and "proximity to a convention center." A simple similarity search might retrieve documents about business, but it would likely fail to assemble the specific constellation of facilities and services that a business traveler implicitly requires. This failure to grasp the underlying relational structure of the domain knowledge prevents the model from making the nuanced inferences that define true intelligence.

To address this tripartite of challenges, the brittleness of hierarchical pipelines, the inflexibility of fine-tuning, and the structural blindness of standard retrieval. We propose a new paradigm grounded in the principle of Inference-only Knowledge Augmentation. Our framework, the Dual-KG Multi-task Inference (DKG-MTI) system, is designed to inject deep, structured domain knowledge dynamically at the point of inference, thereby harnessing the formidable reasoning power of modern LLMs without the need for costly and static retraining.

The architectural core of DKG-MTI is its novel dual-knowledge-graph structure. It maintains a comprehensive, static Global Hotel-KG, which serves as a repository of objective, domain-wide knowledge, encoding detailed information about hotel attributes, facilities, services, and the established relationships between them. Crucially, in parallel, it constructs a dynamic and ephemeral User-Specific Intent-KG for each incoming review. This user-centric graph is built on-the-fly, capturing the unique preferences, sentiments, and contextual cues present in that single piece of discourse. It represents the user's subjective worldview as expressed in their own words. The core innovation of our framework lies in the mechanism that bridges these two graphs: semantic alignment. Instead of performing a simple keyword-based retrieval, DKG-MTI computes the structural correspondence between the dynamic User-KG and the static Hotel-KG. This process allows the system to identify the semantic gap where the subtle and often profound discrepancies between what a specific user implicitly desires and what the domain objectively offers. This identified gap, representing the user's unstated needs or unexpected disappointments, becomes a powerful, structured contextual signal that guides the LLM's reasoning process. It moves beyond what was said to what was meant. Armed with this deep, structural context, DKG-MTI employs a unified, single-stage multi-task inference engine.

This paper makes three primary contributions to the Information Systems and NLU literature. First, we introduce and validate a novel inference-only architecture that leverages dual-graph semantic alignment to achieve promising performance, demonstrating that dynamic knowledge injection can supersede static, task-specific fine-tuning. This has profound implications for the economic viability and operational agility of deploying AI systems. Second, we formalize the concept of the semantic gap as a principled mechanism for reasoning about the delta between subjective user needs and objective domain knowledge, offering a new theoretical lens for personalization. Third, we propose a unified multi-task architecture that jointly predicts user type and generates intent, providing an empirically validated blueprint for building more robust, accurate, and scalable intent inference engines. By bridging structured knowledge and LLMs within an adaptive, inference-only paradigm, DKG-MTI offers a path toward more intelligent and responsive digital ecosystems in the tourism industry and beyond.

The remainder of this paper is organized as follows: Related work reviews related work in intent inference and knowledge-grounded NLP. Methodology details the architecture of the DKG-MTI framework. Experiments describes our experimental setup and presents the results of our evaluation. Finally, Conclusion discusses the implications of our findings and concludes the paper.

## Related Work

This chapter situates our work within the broader landscape of academic research, drawing connections to established and emerging fields. We begin by reviewing the evolution of user intent inference, from traditional models to modern neural architectures. We then delve into the critical area of knowledge-grounded natural language processing, specifically focusing on the prevailing paradigms for integrating large language models with knowledge graphs. This review allows us to identify a crucial research gap concerning the dynamic, structure-aware alignment of subjective and objective knowledge, a gap that our DKG-MTI framework is precisely designed to address. By contextualizing our contributions, we highlight the novelty and significance of our proposed methodology.

### User Intent Inference and Modeling

The accurate inference and modeling of user intent stand as a cornerstone for the next generation of intelligent information systems, driving significant advancements in personalized recommendation, conversational AI, and human-computer interaction (Sun & Duan, 2025). In the modern digital ecosystem, the focus has shifted from processing explicit user queries to understanding the complex, often implicit, goals and motivations latent within user-generated content. Discerning whether a user is "planning a honeymoon" versus "arranging a last-minute business trip" from their review history, for instance, requires a level of semantic comprehension that goes far beyond surface-level text analysis (Ding et al., 2024). It is this deep understanding that enables systems to deliver truly adaptive and context-aware services that anticipate user needs rather than merely reacting to them.

The historical trajectory of user intent modeling began with methods that, while foundational, were limited in their semantic capabilities. Early approaches primarily relied on keyword matching and rule-based systems, which were brittle and struggled with the lexical diversity of natural language (e.g., synonyms, polysemy). Statistical methods like Latent Dirichlet Allocation (LDA) for topic modeling offered a more abstracted view by identifying thematic clusters within documents (Blei et al., 2003). However, while LDA could reveal that a review discusses "food," "price," and "service," it could not capture the specific, relational intent, such as a user's search for a "cheap restaurant with good service." The subsequent wave of traditional machine learning techniques, including Support Vector Machines (SVMs) and Naive Bayes classifiers, marked an improvement by leveraging hand-crafted features like n-grams and TF-IDF vectors (Joachims, 1998). Nevertheless, this paradigm was heavily reliant on extensive and labor-intensive feature engineering and remained ill-equipped to handle the deep contextual dependencies inherent in complex user expressions (Bengio et al., 2013).

The advent of deep learning, and particularly the rise of neural networks, revolutionized the field of Natural Language Understanding (NLU). Architectures such as Recurrent Neural Networks (RNNs), and their more sophisticated variants, Long Short-Term Memory (LSTM) and Gated Recurrent Unit (GRU) networks were among the first to effectively model the sequential nature of text, allowing them to capture word order and short-term contextual dependencies far better than their predecessors (Liu et al., 2016). This made them highly effective for tasks like dialogue state tracking and turn-by-turn intent classification in conversational agents. The introduction of the Transformer architecture, and specifically the self-attention mechanism, marked another paradigm shift (Vaswani et al., 2017). Pre-trained language models like BERT leveraged this architecture to learn deep, bidirectional contextual representations, enabling a model to understand that the meaning of a word is contingent upon the entire sentence (Devlin et al., 2019), this breakthrough set a new standard on numerous NLU benchmarks and was widely adapted for intent classification tasks.

More recently, the research community has increasingly shifted from discriminative models, which classify text into predefined categories, to large-scale generative models like GPT (Radford et al., 2019). These models are capable not only of classifying intent but of articulating it in rich, natural language, a task closely aligned with the "reverse intent generation" objective of this paper. However, despite their impressive fluency and generalization capabilities, these massive, data-driven models are not a panacea. They often operate as opaque "black boxes" and, when ungrounded, are prone to factual inconsistencies and "hallucinations," particularly in specialized or rapidly evolving domains (Brown et al., 2020). For example, a base LLM may know from statistical co-occurrence that "business trip" is often mentioned with "Wi-Fi," but it lacks the structured, relational knowledge to understand that reliable Wi-Fi is a *prerequisite* for a successful business trip, not just a correlated term. This fundamental limitation—the gap between general linguistic competence and specific, structured domain expertise, underscores the critical need for external knowledge grounding, a challenge that our work directly confronts.

### Integrating Knowledge Graphs with Language Models for Intent Inference

To address the knowledge limitations of large language models, a significant body of research has focused on integrating them with external, structured knowledge sources, most notably knowledge graphs (KGs). The core motivation is to imbue LLMs with factual, domain-specific, and relational knowledge, thereby enhancing their reasoning capabilities, improving factual accuracy, and increasing the explainability of their outputs. Within this domain, several distinct paradigms for KG-LLM integration have emerged, each with its own architectural philosophy, advantages, and drawbacks.

The first major paradigm is based on fine-tuning. In this approach, a general-purpose pre-trained language model is further trained on domain-specific corpora, which can include textual documents or linearized representations of knowledge graph triples. By adjusting the model's parameters, this process aims to inject the domain knowledge directly into the LLM's weights. Parameter-Efficient Fine-Tuning (PEFT) techniques, such as LoRA (Hu et al., 2022), have made this process more computationally feasible. The HII-KG framework is a representative example of this paradigm, where a model is fine-tuned for aspect-based classification and subsequent intent generation (Peng et al., 2026). While effective at adapting a model to a specific domain, this approach suffers from a critical flaw: knowledge fossilization. The learned information becomes static and embedded within the model. In dynamic domains like tourism, where hotel services change and new trends emerge, the model quickly becomes outdated, requiring costly and continuous retraining cycles to stay relevant.

A second, more flexible paradigm is RAG. RAG-based systems decouple the knowledge source from the LLM's parameters. At inference time, they use a retriever module to fetch relevant information from an external knowledge base which can be a collection of documents or a KG, and provide this information as context to the LLM along with the user's query (Gao et al., 2023; Wang et al., 2025). This approach offers significant advantages in terms of knowledge freshness, as the external knowledge base can be updated independently of the LLM (Guu et al., 2020). However, standard RAG implementations often exhibit structural blindness. They typically rely on vector similarity search over a flat collection of text chunks, failing to leverage the rich relational topology of a knowledge graph (Pan et al., 2024). This limits their ability to perform complex, multi-hop reasoning (Edge et al., 2024). Recognizing this, recent works like G-Retriever have begun to explore more sophisticated retrieval strategies directly on graph structures (He et al., 2024). These methods, however, often focus on retrieving subgraphs from a single, monolithic KG rather than aligning knowledge from disparate graph sources, which is a key focus of our work.

This leads to the third and most recent trend: dynamic knowledge construction. This emerging paradigm leverages the powerful generative and NLU capabilities of LLMs to construct knowledge graphs on-the-fly from unstructured or semi-structured text. Frameworks like PROM and Intention KG demonstrate the feasibility of using LLMs to extract entities and relations from text to build personalized or intention-focused graphs in real-time (Maoliniyazi et al., 2024; Bai et al., 2024). This line of research is closely related to the first stage of our DKG-MTI framework, where we dynamically construct a User-Specific Intent-KG $G_U$. However, these works often treat the constructed graph as the final output or use it for direct analysis. They typically do not address the subsequent, crucial step of aligning this newly formed, subjective graph with a comprehensive, objective global knowledge base to identify semantic gaps and enrich the reasoning context. It is precisely in this cross-graph alignment that the primary novelty of our DKG-MTI framework lies. By first dynamically modeling the user's perspective and then structurally aligning it against a global domain ontology, our method bridges the gap between the subjective and the objective, enabling a more nuanced and deeply contextualized form of intent inference that current paradigms do not fully address.

To sum up, our review of the existing literature reveals a distinct research gap at the intersection of user intent modeling and knowledge-grounded LLMs. While current paradigms have made significant strides, they are hampered by a persistent trade-off: fine-tuning-based methods achieve domain specificity at the cost of flexibility and freshness, leading to fossilized knowledge. Conversely, conventional RAG systems offer flexibility but often suffer from structural blindness, failing to leverage the rich relational context of knowledge graphs. Although emerging dynamic graph construction techniques show promise, they have largely focused on building user-specific graphs without addressing the critical challenge of aligning this subjective, ephemeral knowledge with a stable, objective domain ontology. This dissertation addresses this gap by proposing the Dual-KG Multi-task Inference (DKG-MTI) framework. Our work makes a novel contribution by introducing a structure-aware semantic alignment mechanism that bridges the subjective user perspective, captured in a dynamically generated User-KG, with the objective domain knowledge of a global Hotel-KG. Instead of merely retrieving information, our framework reasons about the relationship between these two knowledge sources. Furthermore, by employing a unified multi-task inference engine, we directly tackle the error propagation problem inherent in the hierarchical models that are common in this field. Our contribution, therefore, is not just an alternative architecture but a new methodological approach to inference-only knowledge augmentation that is simultaneously structured, dynamic, and robust.

# Methodology

The preceding sections have elucidated the inherent limitations of conventional approaches to user intent inference within dynamic domains such as tourism. Specifically, we identified three critical shortcomings: the susceptibility of hierarchical pipelines to error propagation, the static and computationally intensive nature of parameter-efficient fine-tuning (PEFT), and the structural blindness of standard retrieval-augmented generation (RAG) systems. To address these multifaceted challenges, this paper introduces the Dual-KG Multi-task Inference (DKG-MTI) framework, a novel methodological paradigm grounded in the principle of Inference-only Knowledge Augmentation. The overall architecture of the DKG-MTI framework is illustrated in Figure 1. The framework is engineered to transcend the aforementioned constraints by dynamically injecting structured, domain-specific knowledge into a Large Language Model (LLM) at the point of inference, thereby fostering a more robust, adaptive, and contextually aware understanding of user intent.

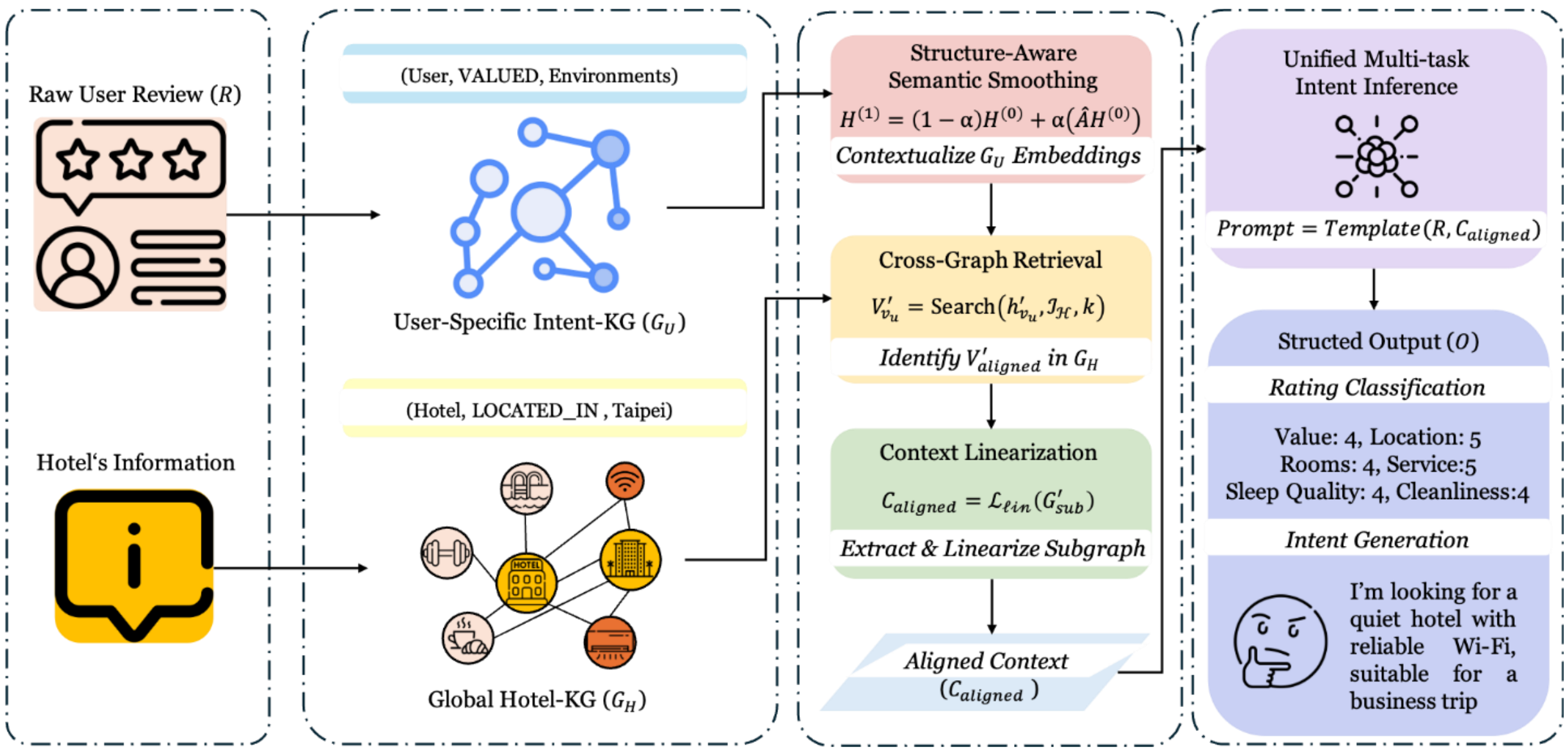

**Figure 1. The Architecture of DKG-MTI**

At its core, DKG-MTI operates through a dynamic interplay between subjective user discourse and objective domain knowledge. Unlike systems that rely on pre-compiled, static knowledge, DKG-MTI actively synthesizes and aligns disparate knowledge sources in real-time. The entire framework can be formally conceptualized as a high-level function, $F$, that maps an input raw user review, $R$, to a structured output object, $O = \{Y_{class}, Y_{intent}\}$. As depicted in the figure, after the inputs, the framework comprises three main phases. First, a Dynamic User-KG Construction component models the user's review into a subjective graph ($G_U$). Second, a Structure-Aware Semantic Alignment mechanism bridges the gap between this user graph and the static Global Hotel-KG ($G_H$), producing a rich, aligned context ($C_{aligned}$). Finally, a Unified Multi-task Intent Inference engine uses both the original review and the aligned context to simultaneously predict aspect ratings ($Y_{class}$) and generate the user's reverse intent ($Y_{intent}$). The following sections will provide a detailed exposition of each component's theoretical underpinnings and algorithmic mechanisms.

### Foundational Knowledge: The Global Hotel-KG

The bedrock of the DKG-MTI framework is the Global Hotel Knowledge Graph ($G_H$), which serves as the immutable, objective, and comprehensive representation of the hotel domain. Unlike the dynamic and ephemeral user-specific knowledge, $G_H$ embodies the collective, structured intelligence about hotels, their attributes, services, and the intricate relationships that define the tourism landscape. This foundational layer is critical for grounding the LLM's reasoning in factual, domain-specific knowledge, thereby mitigating hallucination and enhancing the interpretability of the generated outputs. Formally, $G_H$ is defined as a directed, labeled multigraph $G_H = (V_H, E_H)$, where $V_H$ is the set of nodes and $E_H$ is the set of edges. The ontological structure of $G_H$ is designed to capture the multifaceted nature of hotel-related information, encompassing a rich variety of entity and relation types. As Figure 2 shows, the platform provide various information of the hotel, we transform the information into a graph nodes and edges. The primary entity types include *Hotel* nodes, which represent individual establishments; *Facility* nodes, denoting amenities like "Swimming Pool" or "Free Wi-Fi"; *RoomType* nodes, specifying categories such as "Suite" or "Family Room"; *Location* nodes for geographical areas; *TravelerType* nodes to categorize user profiles like "Business Traveler"; and *Aspect* nodes for standard evaluation criteria such as "Value" or "Cleanliness". These entities are interconnected by a set of typed, directed relations, such as *HAS_FACILITY*, *LOCATED_IN*, and the crucial *PREFERS* relation, which models the probabilistic affinity of a *TravelerType* for other entities. This formal definition ensures a consistent and machine-readable representation of domain knowledge, which is paramount for the subsequent graph-based reasoning.

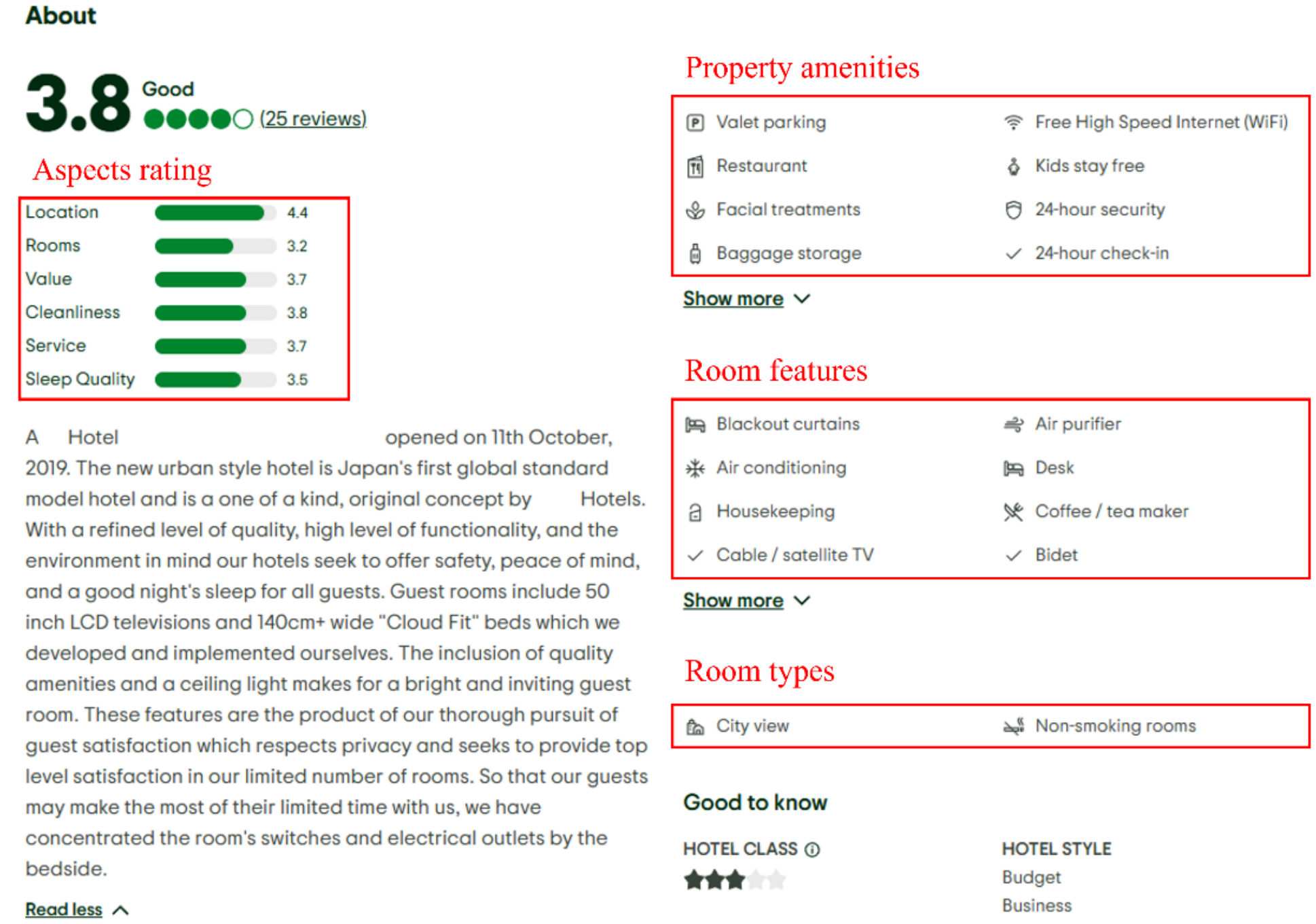

**Figure 2. The information of the hotel.** **(Source from TripAdivsor platform.)**

The construction of $G_H$ is a systematic, multi-stage process that leverages large-scale, publicly available hotel review datasets and structured hotel metadata. The subsequent relation extraction phase establishes connections between these entities using a hybrid approach. This includes using predefined linguistic rules and dependency parsing for explicit relations, employing specialized LLMs for identifying more complex and implicit connections, and applying statistical co-occurrence analysis to infer weaker, probabilistic relationships. A key aspect of this stage is preference inference, where *PREFERS* relations are established based on aggregated user ratings and sentiment analysis. Finally, the extracted entities and relations are populated into a graph database, which undergoes iterative refinement, including manual curation and conflict resolution, to maintain the accuracy and completeness of the knowledge graph over time.

To enable efficient semantic querying and alignment with the dynamic User-KG, each node $v_h \in V_H$ in the Global Hotel-KG is associated with a dense, fixed-dimensional vector embedding. This vectorization is achieved by applying a robust pre-trained sentence-transformer model, $\mathcal{E}_{\mathcal{H}}$:text $\rightarrow R^d$, to the textual representation of each entity, such as its name and description, producing an embedding $e_{v_h} \in R^d$. These embeddings are then stored in a highly optimized Approximate Nearest Neighbor (ANN) index (Arya et al., 1998), $\mathcal{I}_{\mathcal{H}}$, which utilizes algorithms like Hierarchical Navigable Small World (HNSW) to balance retrieval speed with accuracy (Malkov & Yashunin, 2018). This indexing is critical, as it allows the framework to perform rapid semantic matching of concepts generated from user reviews against the established domain ontology. For any given query vector, $\mathcal{I}_{\mathcal{H}}$ can efficiently return a set of the most similar entities from $G_H$ based on a chosen similarity metric, thereby forming the crucial basis for the cross-graph alignment performed in the subsequent stages of the pipeline.

### User Intent Modeling via Dynamic Graph Construction

While the $G_H$ provides the objective, factual backbone for our framework, understanding a user's specific intent requires capturing their subjective perspective as articulated in their review. The first active stage of the DKG-MTI pipeline is therefore the real-time construction of a User-Specific Intent Knowledge Graph ($G_U$). This graph is ephemeral, lightweight, and tailored to a single user review, serving as a structured representation of that user's unique discourse, including their expressed preferences, sentiments, and the semantic relationships between the concepts they mention. This process transforms the unstructured, linear sequence of text into a structured, relational format amenable to graph-based analysis.

Formally, given a raw user review, $R$, represented as a sequence of tokens, this stage applies a specialized function, $\mathcal{F}_{extract}$, to generate a set of semantic triples, $T_U$. This function is operationalized by a large language model (LLM) guided by a carefully designed few-shot prompt. The prompt instructs the LLM to act as an expert knowledge engineer, identifying and extracting salient entities and their relationships from the review text. The extraction process is defined as:

$$T_U = \mathcal{F}_{extract}(R) = \{(s_i, p_i, o_i)\}_{i=1}^{k}.$$

Each resulting triple, $(s_i, p_i, o_i)$, consists of a subject, a predicate, and an object. To maintain a user-centric perspective, the subject $s_i$ is almost always normalized to the canonical entity User. The predicate $p_i$ captures the nature of the user's sentiment or action (e.g., *LIKED*, *DISLIKED*, *MENTIONED*, *VALUED*), and the object $o_i$ represents the entity or concept being discussed (e.g., "Small Room", "Quiet Environment"). For instance, the sentence "The room was surprisingly small, but I really valued the quiet environment" would be deconstructed into two triples: (User, DISLIKED, Small Room) and (User, VALUED, Quiet Environment).

From the extracted set of triples, $T_U$, the User-Specific Intent-KG, $G_U = (V_U, E_U)$, is constructed. The set of nodes, $V_U$, is formed by the union of all unique subjects and objects present in the triples: $V_U = \{s_i\} \cup \{o_i\}$. The set of directed edges, $E_U$, is derived from the predicates, connecting the corresponding subject and object nodes. This transformation from a flat set of triples to a graph structure is crucial, as it establishes an explicit topology of the user's expressed thoughts. This topology, even if simple, provides the structural information necessary for the subsequent graph smoothing stage, where the semantic meaning of each node will be refined by the context of its neighbors. The resulting graph, $G_U$, thus serves as a personalized, structured mental map of the user's review, ready for alignment against the objective domain knowledge of $G_H$.

### Cross-Graph Alignment via Structure-Aware Semantic Smoothing

The technical crux of the DKG-MTI framework lies in this stage, where we bridge the semantic chasm between the subjective, ephemeral User-KG ($G_U$) and the objective, static Global Hotel-KG ($G_H$). A naive approach might involve directly matching entities from $G_U$ to $G_H$ based on lexical similarity. However, such a method is inherently brittle and fails to account for the profound impact of context on meaning. For example, the entity "service" can have vastly different implications depending on whether it is discussed in the context of a "restaurant" or "room cleaning." To address this, we introduce a structure-aware alignment mechanism that leverages principles from Graph Signal Processing to refine entity representations before performing the cross-graph mapping. This process ensures that the alignment is not merely lexical but deeply contextual and semantic.

The alignment process begins with the generation of initial, context-agnostic feature representations for the nodes in the User-KG. We employ a pre-trained sentence-transformer model, $\mathcal{E}_U$:string $\rightarrow R^d$, to map the textual content of each node $v_u \in V_U$ into a $d$-dimensional vector space. This initial embedding, $h_{v_u}^{(0)} = \mathcal{E}_U(\text{text}(v_u))$, captures the general meaning of the entity. These initial embeddings for all nodes in $G_U$ are stacked to form the initial node feature matrix, $H^{(0)} \in R^{|V_U| \times d}$.

The core of our structure-aware alignment is the application of a graph smoothing operation, which functions as a low-pass filter on the graph's signal, effectively allowing nodes to absorb semantic context from their immediate neighbors. This is a non-parametric, inference-only application of a Graph Convolutional Network (GCN) layer. Let $A$ be the adjacency matrix of $G_U$ (where $A_{ij} = 1$ if there is an edge from node $i$ to node $j$) and $D$ be its diagonal degree matrix, with $D_{ii} = \sum_j A_{ij}$. We compute the row-normalized adjacency matrix $\hat{A} = D^{-1}A$. The smoothed, context-aware feature matrix, $H^{(1)}$, is then calculated as a weighted average of the initial features and the neighbor-aggregated features:

$$H^{(1)} = (1 - \alpha)H^{(0)} + \alpha(\hat{A}H^{(0)}).$$

Here, α ∈ [0,1] is a crucial hyperparameter that controls the degree of smoothing. A higher α places more weight on the neighborhood context, while a lower α retains more of the node's original, isolated meaning. This operation mathematically infuses the representation of each node with the semantic essence of its neighbors. For instance, the vector for "service" connected to "restaurant" in $G_U$ will be shifted in the embedding space towards concepts like "dining service" found in $G_H$, thus disambiguating its meaning.

Armed with these contextually-refined embeddings, $H^{(1)}$, we proceed to the cross-graph retrieval step. For each node $v_u \in V_U$, we extract its smoothed embedding, $h'_{v_u}$ (the corresponding row from $H^{(1)}$), and use it as a query vector against the pre-computed ANN index $\mathcal{I}_{\mathcal{H}}$ of the Global Hotel-KG. This semantic search retrieves the top-$k$ most similar entities from $G_H$, denoted as $V'_{v_u} = \text{Search}(h'_{v_u}, \mathcal{I}_{\mathcal{H}}, k)$. The union of these retrieved sets, $V'_{aligned} = \bigcup_{v_u \in V_U} V'_{v_u}$, forms a set of anchor nodes in $G_H$ that are semantically aligned with the user's discourse.

Finally, to create a coherent contextual input for the downstream LLM, we perform context linearization. We extract a rich subgraph, $G'_{sub}$, from $G_H$ by performing a multi-hop neighborhood exploration (e.g., 2-hops) around the aligned anchor nodes $V'_{aligned}$. This subgraph, which contains not only the aligned nodes but also their relevant neighbors and relationships in the global graph, represents the most pertinent domain knowledge for understanding the specific user's review. This subgraph is then converted by a linearization function, $\mathcal{L}_{\ell in}$, into a fluent natural language string, $C_{aligned}$. This string, $C_{aligned} = \mathcal{L}_{\ell in}(G'_{sub})$, serves as the structured knowledge augmentation for the ultimate inference stage.

### Unified Multi-task Intent Inference

The culmination of the DKG-MTI pipeline is the final inference stage, which represents a significant departure from conventional hierarchical models. Having meticulously constructed a rich, contextualized knowledge base through dynamic graph generation and structure-aware alignment, this stage leverages a single, powerful Large Language Model, denoted $\mathcal{L}_{\mathcal{MTI}}$, to perform two distinct yet interdependent tasks in a unified, non-autoregressive manner. This unified approach, a core tenet of our framework, is designed explicitly to overcome the error propagation issues inherent in multi-stage pipelines. By compelling the model to consider both classification and generation tasks simultaneously, we enforce a strong contextual consistency, ensuring that the generated intent is logically coherent with the inferred user profile or aspect ratings.

The input to the final reasoner, $\mathcal{L}_{\mathcal{MTI}}$, is a carefully constructed prompt, $P_{in}$, which is a concatenation of the original, unprocessed user review, $R$, and the linearized, aligned contextual knowledge string, $C_{aligned}$, derived from the previous stage. The prompt is structured using a template, $\mathcal{T}$, that clearly delineates the roles of the different information sources and specifies the required multi-task output format. The complete input prompt is thus formulated as:

$$P_{in} = \mathcal{T}(R, C_{aligned}).$$

This prompt design ensures that the LLM has access to both the user's original, nuanced language in $R$ and the structured, factual grounding provided by $C_{aligned}$. The model is thereby empowered to cross-reference the user's statements against the domain knowledge, enabling a more sophisticated and factually consistent reasoning process.

Upon receiving the input prompt $P_{in}$, the LLM generates a single, structured output object, $O$, which encapsulates the entire result of the inference process. This is a generative task where the model is instructed to produce a well-formed JSON object adhering to a strict schema. The generation function is represented as:

$$O = \mathcal{L}_{\mathcal{MTI}}(P_{in}).$$

The output object, $O = \{Y_{class}, Y_{intent}\}$, contains two key components. The first, $Y_{class}$, represents the results of the classification task: a multi-dimensional vector of predictions for the six standard hotel aspects (e.g., `"classification": {"Value": 4, "Location": 5, "Rooms": 3, "Service": 5, "Sleep Quality": 4, "Cleanliness": 5}`). The second component, $Y_{intent}$, is the output of the generative task: a concise, first-person natural language statement that synthesizes the user's underlying goals and preferences, known as the reverse intent. For example, `"intent": "I am looking for a quiet hotel with reliable Wi-Fi, suitable for a business trip."`. By generating these two outputs within the same structured object in a single pass, the model is implicitly trained (or prompted, in a zero-shot scenario) to ensure that the generated intent $Y_{intent}$ is a logical consequence of the classification results $Y_{class}$ and the provided context. This unified, multi-task architecture stands in stark contrast to pipelined approaches and represents a more robust and holistic method for user intent inference.

# Experiments

To empirically validate the efficacy and robustness of the proposed DKG-MTI framework, we conducted a series of comprehensive experiments. This section is structured to systematically evaluate the core contributions of our methodology. First, we will detail the experimental setup, including the dataset, the evaluation metrics designed to assess both classification accuracy and generative quality, and the specifics of our implementation. Following this, we present the main results, where our full DKG-MTI framework is benchmarked against a suite of carefully selected baseline models, ranging from zero-shot large language models to various forms of retrieval-augmented and hierarchical systems. Subsequently, we conduct a thorough ablation study to dissect the DKG-MTI framework itself, quantifying the performance contribution of each of its key architectural components. Finally, we provide a qualitative analysis through case studies to offer intuitive insights into the model's behavior and demonstrate its superiority in handling complex, real-world scenarios.

## Experimental Setup

The dataset employed in this study was curated from publicly available English-language reviews on TripAdvisor, focusing on 745 well-known hotels. From an initial pool of 10,000 reviews, we randomly sampled 1,000 reviews to form our experimental testbed. Each review in this subset is accompanied by a rich set of ground-truth labels, including authentic, user-provided ratings for six aspects (Value, Location, Rooms, Service, Sleep Quality, Cleanliness) and the traveler type (e.g., Business, Family). A crucial component of our dataset is the ground-truth reverse intent, which was produced through a rigorous manual annotation process. Human annotators were tasked with reading each review and formulating a concise, first-person statement summarizing the likely pre-booking intent of the reviewer. This provides a high-quality benchmark for our generative task.

Our evaluation is twofold, addressing both the classification and generation facets of the multi-task output. For the classification task (aspect rating prediction), we report the macro-averaged F1-Score (F1) to provide a balanced view of performance across all rating classes. For the intent generation task, we employ three types of metrics. First, we use BLEU and ROUGE-L to measure the lexical overlap between the generated intent and the human-annotated ground truth, providing a measure of surface-level similarity. Second, to capture the more critical aspect of semantic correctness and coherence, we utilize an LLM-as-a-Judge approach (Zheng et al., 2023). Specifically, we employ *prometheus-7b-v2.0* to score the quality of the generated intent on a scale of 1 to 5 based on its accuracy and semantic alignment with the ground truth, providing a robust measure of generative quality (Kim et al., 2024).

To comprehensively assess the performance of our DKG-MTI framework, we compare it against a diverse set of baselines. These include powerful, proprietary API-based models such as OpenAI's *gpt-4o-mini* and Google's *gemini-1.5-pro-latest*, as well as a open-source model, *Mistral-7B-Instruct-v0.2* (Jiang et al., 2023), all evaluated in a zero-shot setting. Furthermore, we present results from the ablation studies detailed in Section 4.3, which also serve as critical points of comparison. Our primary DKG-MTI model utilizes *Llama-3.1-8B-Instruct* (Grattafiori et al., 2024) as its core reasoning engine. For all graph-based vector operations, entity embeddings are generated using the sentence-transformers model (Reimers & Gurevych, 2019). The graph data structures are managed using a Neo4j database. Based on analysis of our framework, key hyperparameters were set as follows: the graph smoothing operation performs a single-pass, equal-weighted averaging of a node's features with its immediate neighbors, conceptually equivalent to setting the smoothing factor $\alpha$ to 0.5 in a simple two-node case (Li et al., 2018). The cross-graph alignment retrieves the single best match, meaning the hyperparameter for top-$k$ nearest neighbor search is set to $k = 1$. The subsequent context extraction from the Global Hotel-KG explores the immediate 1-hop neighborhood of the aligned node. All experiments were conducted on a workstation equipped with an NVIDIA GeForce RTX 5090 GPU with 32GB of VRAM and 128GB of system RAM.

### Main Results

This section presents the core findings of our empirical evaluation, comparing the performance of our proposed DKG-MTI framework against the 6 baselines. The results, summarized in Table 1, provide a comprehensive and multi-faceted overview of each model's capabilities. We report the F1-Score for each of the six classification aspects individually, as well as the macro-average, to allow for a granular analysis. For the generative task, we report both lexical and semantic metrics to give a complete picture of intent inference quality. The primary goal is to demonstrate the tangible benefits of our structured, dual-graph approach compared to both knowledge-agnostic LLMs and simpler retrieval-augmented methods.

The empirical evaluation of the DKG-MTI framework against a comprehensive suite of baselines reveals its superior performance across both multi-label aspect classification and generative intent inference tasks. As detailed in Table 1, which presents the F1-Scores for individual aspects and their macro-average, DKG-MTI consistently outperforms all compared methods in the aspect rating classification. Notably, our framework achieves a macro-averaged F1-Score of 0.83, surpassing the strongest proprietary API model, Gemini-1.5-Pro (0.76), and the hierarchical HII-KG (0.79) by significant margins. This robust performance is evident across all six aspects, with DKG-MTI demonstrating particular strength in *Value* (0.88), *Location* (0.91), and *Cleanliness* (0.88), indicating its enhanced capability to accurately discern fine-grained user sentiments even for nuanced attributes. The substantial improvement over knowledge-agnostic LLMs like Llama-3.1-8B (0.70) and Mistral-7B (0.68) underscores the critical role of knowledge augmentation in grounding LLM reasoning for domain-specific classification tasks. Furthermore, DKG-MTI's advantage over Naive RAG (0.78) and HII-KG highlights the efficacy of our structure-aware semantic alignment and unified multi-task architecture in leveraging knowledge graphs more effectively than simpler retrieval or hierarchical approaches.

The benefits of the DKG-MTI framework are even more pronounced in the generative intent inference task, as evidenced by the results in Table 2. Our model achieves the highest scores across all generative metrics: BLEU (0.417), ROUGE-L (0.497), and, most critically, the LLM-as-a-Judge score (4.18). The LLM-as-a-Judge metric, which provides a semantic assessment of generated intent quality, demonstrates DKG-MTI's ability to produce outputs that are not only lexically similar to human-annotated ground truths but also semantically accurate, coherent, and contextually relevant. DKG-MTI's LLM-as-a-Judge score of 4.18 significantly surpasses that of Gemini-1.5-Pro (4.05) and GPT-4o-mini (3.96), which are considered state-of-the-art proprietary models. This outcome is particularly noteworthy as it suggests that our inference-only knowledge augmentation strategy, coupled with the unified multi-task reasoning, enables a smaller, open-source LLM (Llama-3.1-8B as the core) to generate higher-quality, more semantically aligned intent statements than even the most advanced general-purpose LLMs operating in a zero-shot capacity. The improvements over Naive RAG (3.65) and HII-KG (3.81) further validate the necessity of our sophisticated cross-graph alignment and the advantages of a unified inference paradigm in mitigating error propagation and ensuring contextual consistency in generative tasks. These results collectively affirm that DKG-MTI effectively bridges the "semantic gap" between user discourse and domain knowledge, leading to a more profound and actionable understanding of user intent.

**Table 1. Overall performance comparison of DKG-MTI in rating classification task.**

| Model | Val. | Loc. | Rooms | Svc. | Slp. | Cln. | Avg. |
|---|---|---|---|---|---|---|---|
| Llama-3.1-8B | 0.72 | 0.78 | 0.66 | 0.69 | 0.60 | 0.75 | 0.70 |
| Mistral-7B | 0.70 | 0.76 | 0.64 | 0.67 | 0.58 | 0.73 | 0.68 |
| GPT-4o-mini | 0.76 | 0.82 | 0.70 | 0.73 | 0.63 | 0.80 | 0.74 |
| Gemini-1.5-Pro | 0.79 | 0.85 | 0.72 | 0.75 | 0.65 | 0.82 | 0.76 |
| Naive RAG | 0.81 | 0.87 | 0.74 | 0.77 | 0.66 | 0.84 | 0.78 |
| HII-KG | 0.83 | 0.88 | 0.73 | 0.77 | 0.65 | 0.85 | 0.79 |
| DKG-MTI (Ours) | **0.88** | **0.91** | **0.80** | **0.81** | **0.70** | **0.88** | **0.83** |

**Table 2. Overall performance comparison of DKG-MTI in intent generation task**

| Model | BLEU | ROUGE-L | LLM-as-a-Judge |
|---|---|---|---|
| Llama-3.1-8B | 0.255 | 0.430 | 3.35 |
| Mistral-7B | 0.242 | 0.421 | 3.28 |
| GPT-4o-mini | 0.336 | 0.478 | 3.96 |
| Gemini-1.5-Pro | 0.352 | 0.484 | 4.05 |
| Naive RAG | 0.300 | 0.455 | 3.65 |
| HII-KG | 0.322 | 0.471 | 3.81 |
| DKG-MTI (Ours) | **0.417** | **0.497** | **4.18** |

## Ablation Study

To dissect the internal mechanics of our DKG-MTI framework and quantify the contribution of each core component, we conducted a series of ablation studies. By systematically removing or simplifying key parts of our pipeline, we can isolate their impact on the final performance. This analysis is crucial for validating our central architectural claims: the necessity of grounding in a global knowledge base, the superiority of structure-aware alignment over simple semantic matching, and the robustness of a unified inference paradigm. The results of these studies are presented in Figure 3, Figure 4.

The most significant performance degradation is observed in the "w/o Global-KG ($G_H$)" setting, where the framework is deprived of the static, objective domain knowledge. In this scenario, the F1-Avg score for classification drops sharply from 83.0% to 80.6%, and the LLM-as-a-Judge score for generation falls from 4.18 to 4.02. This substantial decline underscores the foundational importance of grounding the model's reasoning in a comprehensive knowledge base. Without access to $G_H$, the model must rely solely on the information present in the ephemeral User-KG, leading to a diminished capacity for factual verification and domain-specific reasoning, which impacts both classification accuracy and the semantic quality of the generated intent. The second ablation, "w/o Smoothing," directly tests the efficacy of our proposed structure-aware semantic smoothing mechanism. By removing the graph smoothing step and relying on the initial, context-agnostic node embeddings for cross-graph alignment, the F1-Avg score decreases to 81.8% and the LLM-as-a-Judge score falls to 4.09. While less severe than the complete removal of the global KG, this drop is highly significant. It provides direct evidence that our graph smoothing operation is critical for disambiguating entities within the user's discourse. By infusing node representations with local context from the User-KG, the smoothing process enables a more precise and contextually relevant alignment with the Global Hotel-KG, thereby improving the quality of the knowledge provided to the final reasoner.

Finally, the "w/o Unified Inference" experiment evaluates our unified multi-task architecture against a traditional hierarchical approach. In this setting, where classification is performed first and its output is then fed into a subsequent generation step, the F1-Avg score drops to 81.2% and the LLM-as-a-Judge score declines to 4.06. This demonstrates the robustness of the unified inference paradigm. By compelling the model to perform both tasks simultaneously, we mitigate the risk of error propagation inherent in pipelined models and enforce a stronger contextual consistency between the classification and generation outputs. The performance degradation, though modest, confirms that a unified approach leads to more coherent and reliable results. Collectively, these ablation studies validate the thoughtful design of the DKG-MTI framework, confirming that each component—the global knowledge base, the structure-aware smoothing, and the unified inference engine, which plays an indispensable role in achieving superior performance.

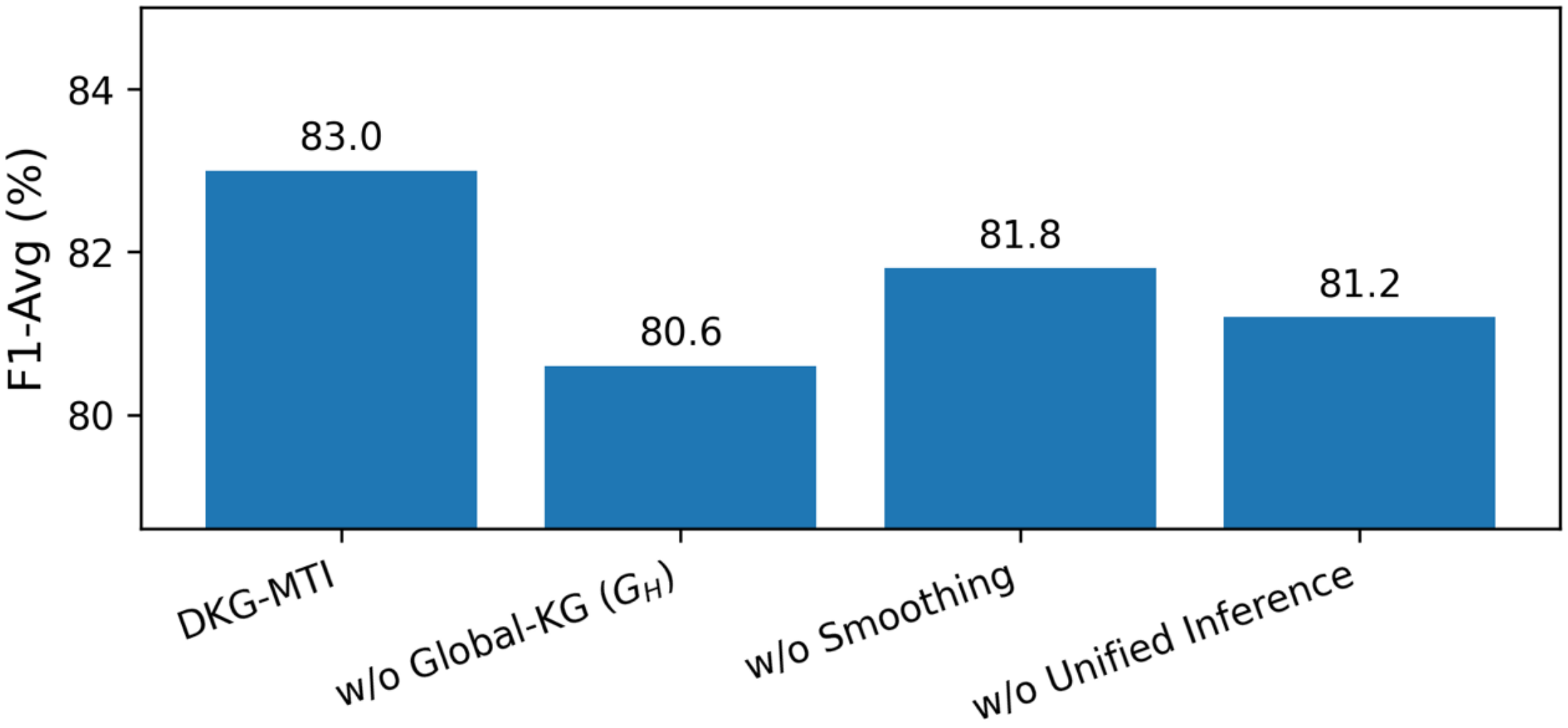

**Figure 3. Ablation study results on the aspect classification task.**

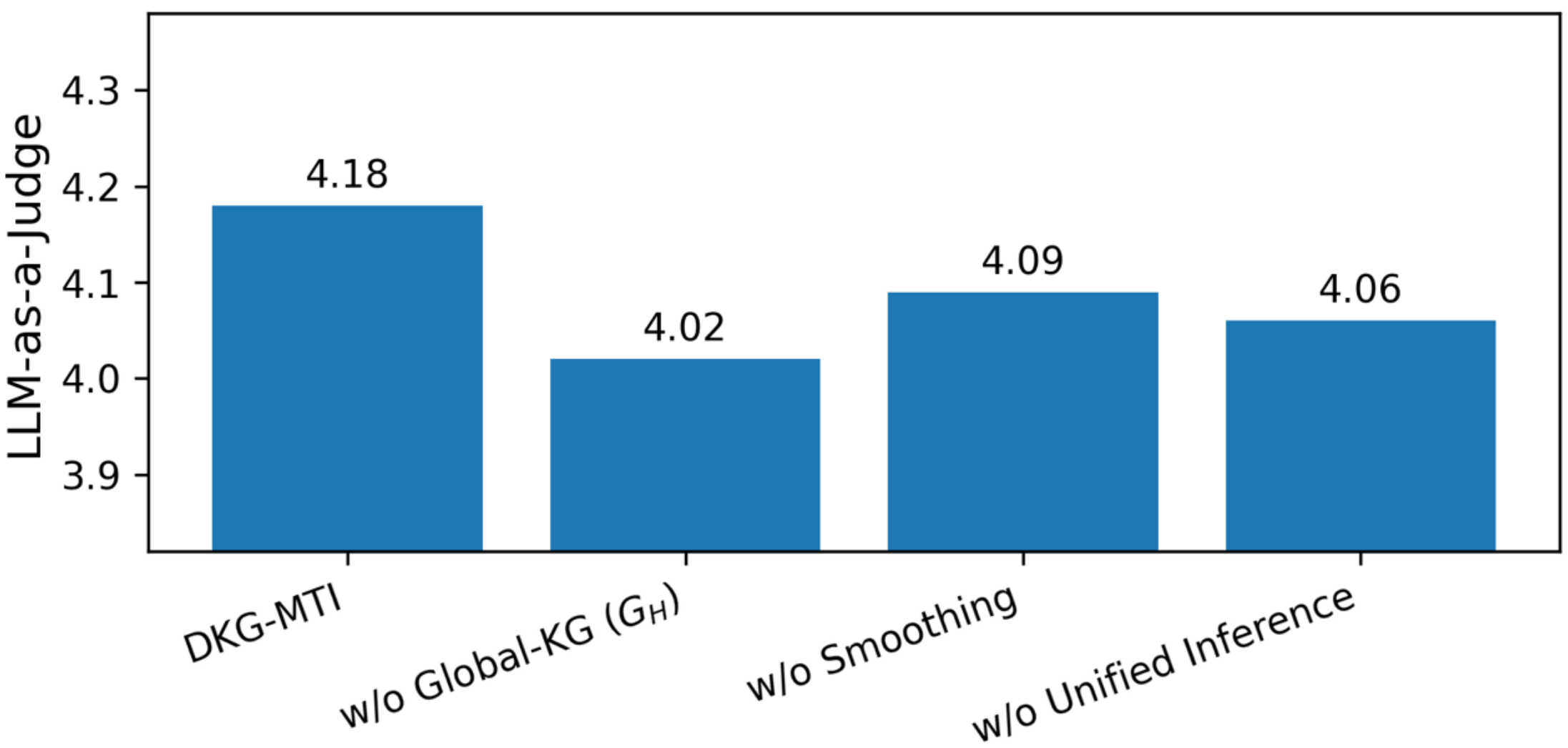

**Figure 4. Ablation study results on the intent generation task.**

**Qualitative Analysis & Case Study**

Beyond quantitative metrics, a qualitative analysis of model outputs provides invaluable insight into the practical advantages of our framework. To this end, we present a case study that illustrates the nuanced reasoning capabilities of DKG-MTI in contrast to key baseline models. We selected a representative user review that contains a mix of explicit and implicit preferences, making it a challenging test case for intent inference.

The outputs in Figure 5 clearly demonstrate the superiority of the DKG-MTI framework. The zero-shot model correctly identifies the location and breakfast but completely misses the crucial context of a business trip, failing to capture the user's primary needs related to work (Wi-Fi) and atmosphere (quiet). The Naive RAG model, lacking structural understanding, performs even worse; it simply lists entities mentioned in the review or retrieved from a generic context, resulting in a nonsensical and unhelpful intent statement.

Most tellingly, the HII-KG model exemplifies the risk of error propagation. It appears to have misclassified the user as a "leisure traveler" in its initial step, perhaps due to the mention of the "hotel bar." This initial error leads it to generate a completely contradictory intent, focusing on a "lively bar" when the user explicitly desired the opposite.

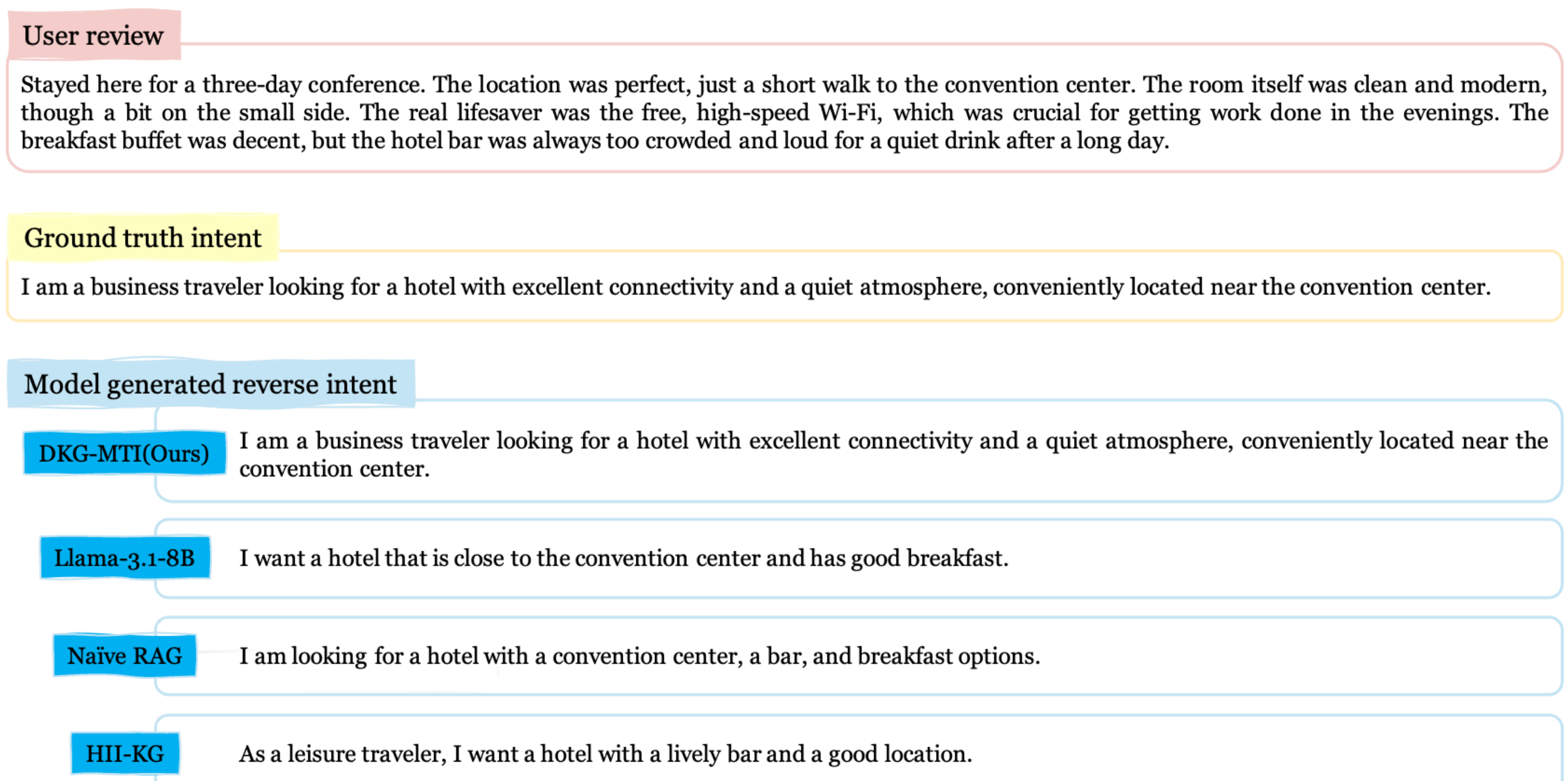

**Figure 5. The comparison of generated reverse intents for a sample user review.**

In stark contrast, our DKG-MTI model successfully synthesizes all key elements. It correctly infers the user's business context, identifies the critical importance of "fast, reliable Wi-Fi" and a "quiet environment," and links them to the explicit need for a location near the "convention center." This demonstrates our framework's ability to leverage structured knowledge to resolve ambiguity, understand implicit needs, and produce a holistically accurate and useful representation of user intent.

## Conclusion

In this paper, we introduced the Dual-KG Multi-task Inference (DKG-MTI) framework, a novel paradigm for user intent inference that addresses the critical shortcomings of contemporary methods. By eschewing static fine-tuning in favor of dynamic, inference-only knowledge augmentation, our framework offers a more agile, scalable, and economically viable solution for real-world information systems. We have demonstrated that the core innovations of our approach—the dual-graph architecture combining a global knowledge base with a real-time, user-specific graph; the structure-aware semantic smoothing mechanism for resolving ambiguity; and the unified multi-task inference engine for ensuring output consistency—collectively contribute to a more robust and accurate understanding of user intent. Our comprehensive experiments, including comparisons against strong baselines and detailed ablation studies, empirically validate the superiority of DKG-MTI, particularly in its ability to produce semantically coherent and factually grounded outputs.

The contributions of this work are threefold. First, we presented a validated, inference-only architecture that effectively injects structured knowledge into LLMs, proving its efficacy as an alternative to task-specific fine-tuning. Second, we formalized and demonstrated the value of structure-aware alignment, showing that considering the topology of user-generated concepts is crucial for accurate contextualization. Third, our unified multi-task design provides a robust blueprint for mitigating error propagation, a persistent challenge in hierarchical information processing pipelines.

Looking ahead, several promising avenues for future research emerge from this work. One immediate direction is to enhance the graph construction and alignment process. The current implementation of graph smoothing is non-parametric; future work could explore lightweight, parametric graph neural network layers that could be trained to optimize the smoothing process for specific domains. Another promising extension is to incorporate temporal dynamics into the Global Hotel-KG, allowing it to model evolving trends, seasonality, and the changing attributes of entities over time. Finally, while our framework excels at inference, exploring its application in a conversational AI or interactive recommendation setting, where the User-KG could be updated and re-aligned in real-time based on dialogue turns, presents an exciting and impactful direction for future research. By continuing to bridge the gap between structured knowledge and the fluid reasoning of large language models, we can pave the way for more intelligent, adaptive, and genuinely personal information systems.